\pdfoutput=1
\documentclass[10pt, a4paper]{article}
\usepackage{lrec2016}
\usepackage[round]{natbib}
\usepackage{graphicx}
\usepackage{epstopdf}
\usepackage[utf8]{inputenc}
\usepackage[T1]{fontenc}
\usepackage[english]{babel}
\usepackage{csquotes}
\usepackage{xstring}
\usepackage{xcolor}
\usepackage{multirow}
\usepackage{subcaption}
\usepackage{amsmath}
\usepackage{nicefrac}
\usepackage{hyperref}

\def\F1{F$_1$} 

\newcommand{\cites}[1]{\citeauthor{#1}'s\ (\citeyear{#1})}
\newcommand{\Section}[1]{Section~\hyperref[sec:#1]{\StrGobbleRight{\getrefnumber{sec:#1}}{1}}}
\newcommand{\Sections}[2]{Sections~\hyperref[sec:#1]{\StrGobbleRight{\getrefnumber{sec:#1}}{1}} and~\hyperref[sec:#2]{\StrGobbleRight{\getrefnumber{sec:#2}}{1}}}
\newcommand{\Table}[1]{Table~\ref{tab:#1}}

\title{Automatic TM Cleaning through MT and POS Tagging: \\Autodesk's Submission to the NLP4TM 2016 Shared Task}

\name{Alena Zwahlen, Olivier Carnal, Samuel Läubli}

\address{Autodesk Development S.à.r.l. \\
         Rue du Puits-Godet 6, 2000 Neuchâtel, Switzerland \\
         \{alena.zwahlen, olivier.carnal, samuel.laubli\}@autodesk.com\\}

\abstract{
We describe a machine learning based method to identify incorrect entries in translation memories. It extends previous work by \citet{Barbu2015} through incorporating recall-based machine translation and part-of-speech-tagging features. Our system ranked first in the Binary Classification (II) task for two out of three language pairs: English--Italian and English--Spanish.
\\ 
\newline 
}

\begin{document}

\maketitleabstract

\section{Introduction}
\label{sec:Introduction}

Autodesk has accumulated more than 40 million professionally translated segments over the past 17 years. These translation units (TUs) mainly stem from user interfaces and documentation of software products localized into 32 languages. As we are now unifying and centralizing all translations in a single repository, it is high time to sort out duplicate, outdated, and erroneous TUs. Exploring methods to handle the latter -- clearly more challenging than removing duplicate and outdated material -- motivated us to participate in the First Shared Task on Translation Memory Cleaning \citep{Barbu2016}. Going forward, we strive to make human translation more efficient (by showing translators less erroneous fuzzy matches) and machine translation more accurate (by reducing noise in training data).

In this paper, we describe our submitted system for distinguishing correct from incorrect TUs. Rather than tailoring it to individual languages, we aimed at a language-independent solution to cover all of the language pairs in this shared task or, looking to the future, Autodesk's production environments. The system is based on previous work by \citet{Barbu2015} and uses language-independent features with language-specific plug-ins,  such as machine translation, part-of-speech tagging, and language classification.

Specifics about previous work are given in the next section. In \Section{Method}, we describe our method and, in \Section{Results}, show how it compares to \cites{Barbu2015} approach as well as other submissions to this shared task. Lastly, we offer preliminary conclusions and outline future work in \Section{Conclusions}.


\section{Background}
\label{sec:Background}

TM cleaning functionality in commercial tools is mostly rule-based, centering around the removal of duplicate entries, ensuring markup validity (e.g., no unclosed tags), or controlling for client or project specific terminology
. Although helpful, these methods fall short of identifying spurious entries that contain language errors or partial translations. With crowd-sourced and automatically constructed TMs in particular, it is also necessary to identify translation units with source and target segments that do not correspond at all \citep[e.g.,][]{Trombetti2009,Tiedemann2012}.

\citet{Barbu2015} has proposed to cast the identification of such incorrect translations as a supervised classification problem. In his work, 1,243 labelled TUs were used to train binary classifiers based on 17 features. The \enquote{most important} of them, according to the author, were \textit{bisegment\_similarity} and \textit{lang\_diff}: the former is defined as the cosine similarity between a target segment and its machine translated source segment, while the latter denotes whether the language codes declared in a translation unit correspond with the codes detected by a language detector. The best classifier, a support vector machine with linear kernel, achieved 82\% precision and 81\% recall on a held-out test set of 309 TUs.

To the best of our knowledge, \citeauthor{Barbu2015} provided the first and so far only research contribution on automatic TM cleaning, which the author himself described as \enquote{a neglected research area} \citep{Barbu2015}. With our participation to this shared task, we seek to extend his work by examining new features based on statistical MT and POS tagging.

As outlined above, comparing machine translated source segments to their actual target segments has proven effective in \cites{Barbu2015} experiments. We propose to complement or replace the similarity function used for this comparison (cosine similarity) by two automatic MT evaluation metrics, Bleu \citep{Papineni2002} and character-based Levenshtein distance, in order to reward higher-order $n$-gram ($n > 1$) and partial word overlaps, respectively. Furthermore, we introduce a recall-based MT feature that takes multiple MT hypotheses ($n$-best translations) of a given source segment into account, based on the assumption that alternative translations of words (such as \enquote{buy} and \enquote{purchase}) or phrases (such as \enquote{despite} and \enquote{in spite of}) should not be punished.

We also experiment with part-of-speech information to identify spurious translation units. With closely related languages in particular, the rationale would be that adjectives (to name an example) in a source segment are likely to be reflected in the corresponding target segment in case of a valid translation. The comparison of POS tags from language-specific tagsets will be based on a mapping to eleven coarse-grained, language-independent grammatical groups \citep{Petrov2011}.

We acknowledge that the use of MT is discouraged by the organizers of this shared task to foster contributions that require less compute power. However, as MT was found to be valuable in previous work (see above) and computational resources are hardly a limiting factor in corporate environments (see \Section{Resources}), we decided not to refrain from including MT-based features in our submissions.

\section{Method}
\label{sec:Method}

\begin{table}
\small
\begin{center}
\setlength{\tabcolsep}{4pt} 
\renewcommand{\arraystretch}{1.3} 
\begin{tabular}{| l | l | l | r|r|r|r |}

      \hline
      \multirow{2}{*}{Language} & \multirow{2}{*}{Domain}   & \multirow{2}{*}{Set}  & \multicolumn{4}{c|}{Translation Units}    \\
                                &                           &                       & 1     & 2     & 3     & Total             \\ \hline\hline
      
      \multirow{2}{*}{en--de}   & \multirow{2}{*}{News}     & Training              & 1,086 & 100   & 210   & 1,396             \\
                                &                           & Evaluation            & 544   & 51    & 105   & 700               \\ \hline

      \multirow{2}{*}{en--es}   & \multirow{2}{*}{Medical}  & Training              & 942   & 128   & 313   & 1,383             \\
                                &                           & Evaluation            & 471   & 65    & 157   & 693               \\ \hline

      \multirow{2}{*}{en--it}   & \multirow{2}{*}{Medical}  & Training              & 872   & 254   & 284   & 1,410             \\
                                &                           & Evaluation            & 437   & 128   & 143   & 708               \\ \hline

\end{tabular}
\caption{Trainig and evaluation data. Classes 1, 2, and 3 denote correct, almost correct, and incorrect translation units, respectively.}
\label{tab:Data}
\end{center}
\end{table}

\begin{table*}[!ht]
\small

\begin{subtable}{1.0\textwidth}
\begin{center}
\begin{tabular}{| l | l | ccc | ccc | ccc |}

      \hline
      \multirow{2}{*}{Features} & \multirow{2}{*}{Algorithm}    & \multicolumn{3}{c|}{en--de}   & \multicolumn{3}{c|}{en--es}   & \multicolumn{3}{c|}{en--it} \\
                                &                               & P     & R     & \F1           & P     & R     & \F1           & P     & R     & \F1         \\
      \hline\hline
      Baseline 1                & Random + Class Distribution   & .77   & .77   & .77           & .67   & .68   & .67           & .70   & .71   & .70         \\
      Baseline 2                & Adapted Church-Gale           & .78   & .77   & .77           & .72   & .71   & .71           & .76   & .74   & .74         \\ \hline
      \citet{Barbu2015}         & SVM (linear kernel)           & .74   & .85   & .78           & .84   & .85   & .83           & .83   & .83   & .79         \\
      \citet{Barbu2015}         & Random Forests                &\bf.80 & .85   &\bf.79         & .88   & .88   & .87           & .87   & .88   & .87         \\
      This work                 & Random Forests                &\bf.80 &\bf.86 &\bf.79         &\bf.89 &\bf.89 &\bf.89         &\bf.90 &\bf.90 &\bf.90       \\
      \hline

\end{tabular}
\caption{Binary Classification (II) task.}
\label{tab:ResultsBinaryTrainingData}
\end{center}
\end{subtable}

\vspace{0.3cm}

\begin{subtable}{1.0\textwidth}
\begin{center}
\begin{tabular}{| l | l | ccc | ccc | ccc |}

      \hline
      \multirow{2}{*}{Features} & \multirow{2}{*}{Algorithm}    & \multicolumn{3}{c|}{en--de}   & \multicolumn{3}{c|}{en--es}   & \multicolumn{3}{c|}{en--it} \\
                                &                               & P     & R     & \F1           & P     & R     & \F1           & P     & R     & \F1         \\
      \hline\hline
      Baseline 1                & Random + Class Distribution   & .63   & .63   & .63           & .53   & .54   & .53           & .45   & .45   & .45         \\
      Baseline 2                & Adapted Church-Gale           & .64   & .63   & .63           & .57   & .57   & .57           & .49   & .48   & .48         \\ \hline
      \citet{Barbu2015}         & SVM (linear kernel)           & .63   & .78   & .69           & .70   & .77   & .73           & .68   & .70   & .61         \\
      \citet{Barbu2015}         & Random Forests                & .72   & .78   & .70           &\bf.80   &\bf.81   &\bf.79           &\bf.75   &\bf.75   &\bf.72         \\
      This work                 & Random Forests                &\bf.77 &\bf.78 &\bf.70         & .74 &\bf.81 & .76         &\bf.75 &\bf.75 & .70       \\
      \hline

\end{tabular}
\caption{Fine-Grained Classification task.}
\label{tab:ResultsFineGrainedTrainingData}
\end{center}
\end{subtable}

\caption{Classification results on training data.}
\end{table*}

\begin{table*}[!ht]
\small

\vspace{3mm}

\begin{subtable}{.5\textwidth}
\begin{center}
\begin{tabular}{| l | cc | cc | cc |}

      \hline
      \multirow{2}{*}{Team} & \multicolumn{2}{c|}{en-de}& \multicolumn{2}{c|}{en-es}&\multicolumn{2}{c|}{en-it} \\
                            & \F1       & c             & \F1       & c             & \F1       & c         \\ \hline \hline
      Baseline 1            & .53       & 534           & .43       & 418           & .50       & 473       \\          
      Baseline 2            & .52       & 532           & .51       & 442           & .56       & 492       \\ \hline
      Autodesk              & .47       & 593           & \bf .81   & \bf 611       & \bf .85   & \bf 644   \\
      Christian Buck-MT     & .66       & 597           & --        & --            & --        & --        \\
      Christian Buck-NMT    & .65       & 594           & --        & --            & --        & --        \\
      FBK-HLTMT             & .49       & 594           & .77       & 596           & .80       & 631       \\
      JUMTTeam              & .58       & 482           & .66       & 493           & .70       & 530       \\
      Lingua Custodia       & .64       & 609           & .78       & 592           & .83       & 634       \\
      Unisa                 & \bf .68   & \bf 618       & .76       & 596           & .77       & 623       \\ \hline

\end{tabular}
\caption{Binary Classification (II) task: averaged \F1-scores and number of correctly classified TUs (c).}
\label{tab:ResultsBinaryFinal}
\end{center}
\end{subtable}
\quad
\begin{subtable}{.5\textwidth}
\begin{center}
\begin{tabular}{| l | c | c | c |}

      \hline
      \multirow{2}{*}{Team} & en-de     & en-es     & en-it     \\
                            & \F1       & \F1       & \F1       \\ \hline \hline
      Baseline 1            & .64       & .48       & .47       \\          
      Baseline 2            & .63       & .52       & .50       \\ \hline
      Autodesk              & .69       & .74       & .68       \\
      Christian Buck-MT     & .77       & --        & --        \\
      Christian Buck-NMT    & .77       & --        & --        \\
      FBK-HLTMT             & .70       & .72       & .66       \\
      Lingua Custodia       & .75       & \bf .79   & .69       \\
      Unisa                 & \bf .80   & .77       & \bf .73   \\ \hline

\end{tabular}
\caption{Fine-Grained Classification task: weighted \F1-scores.}
\label{tab:ResultsFineGrainedFinal}
\end{center}
\end{subtable}

\caption{Official evaluation and ranking results. Winning systems are highlighted in bold type.}
\end{table*}

Our system uses labelled TUs to train classifiers based on language-independent features (see \Section{Features}) with language-specific plug-ins (see \Section{Resources}). The feature extraction pipeline is implemented in Scala (see \Section{Classification}), and our final submission -- geared to distinguish correct or almost correct from incorrect TUs -- is based on a selection of nine features (see \Section{FeatureSelection}).

\subsection{Features}
\label{sec:Features}

We re-implemented the 17 features proposed by \citet[see also \Section{Background}]{Barbu2015}. In addition, we explore

\begin{itemize}
    
    \item \textit{mt\_coverage} \quad
    the percentage of target words contained in the $n$-best machine translations of the source segment. We use $n=20$ in our experiments.
    
    \item \textit{mt\_cfs} \quad
    the character-based Levenshtein distance between target segment and machine translated source segment. We normalise this score such that identical and completely dissimilar segments result in scores of 1.0 and 0.0 respectively, i.e.,
    \begin{equation}\nonumber
        \textit{cfs} = 1 - \frac{\text{Levensthein distance in characters}}{\text{number of characters in longer segment.}}
    \end{equation}
    This score is computed individually for each of the 20-best translation options; the best of these scores instantiates the feature value.
    
    \item \textit{mt\_bleu} \quad
    the BLEU score \citep{Papineni2002} between target segment and machine translated source segment. We employ the sentence-level version of the metric as implemented in Phrasal \citep{Green2014}. As with \textit{mt\_cfs}, individual scores are computed for each of the 20-best translation options; the best score instantiates the feature value.
    
    \item \textit{pos\_sim\_all} \quad
    the cosine similarity between the part-of-speech (POS) tags found in the source and target segment.
    
    \item \textit{pos\_sim\_some} \quad
    the cosine similarity between source and target segment in terms of nouns (\texttt{NOUN}), verbs (\texttt{VERB}), adjectives (\texttt{ADJ}), and pronouns (\texttt{PRON}).
    
    \item \textit{pos\_exact} \quad
    whether or not the POS tag sequence in source and target segment is identical.
    
    \item \textit{language\_detection} \quad
    whether or not a state-of-the-art language classifier confirms the target segment's language declared in the translation unit.
    
    \item \textit{ratio\_words} \quad
    the ratio between number of words in source and target segment.
    
    \item \textit{ratio\_chars} \quad
    the ratio between number of characters in source and target segment. 
    
\end{itemize}

\subsection{Resources}
\label{sec:Resources}

Some of the features described in the previous section require natural language processing (NLP) facilities. For machine translation, we use our in-house systems \citep{PlittMasselot2010,Zhechev2014} based on the Moses SMT framework \citep{Koehn2007}. They are trained on translated software and user manuals from Autodesk products only and chosen for the sake of convenience; we would expect better performance of our MT-based features in conjunction with MT engines geared to the text domains used in this shared task (listed in \Table{Data}). Our engines are integrated into a scalable infrastructure deployed on an elastic compute cloud, allowing high throughput even with large translation memories to be cleaned.

For POS tagging, we rely on \cites{Schmid1995} TreeTagger and its readily available models\footnote{\url{http://www.cis.uni-muenchen.de/~schmid/tools/TreeTagger/}} for English, German, Italian, and Spanish. To make POS tags comparable across these languages, they are mapped\footnote{\url{https://github.com/slavpetrov/universal-pos-tags}} to the Universal Tagset proposed by \citet{Petrov2011}.

Lastly, we use the publicly available Xerox Language Identifier API\footnote{\url{https://open.xerox.com/Services/LanguageIdentifier}} for language detection.

\subsection{Classification}
\label{sec:Classification}

Our feature extraction pipeline, including \cites{Barbu2015} as well as our own features (see \Section{Features}), is implemented in Scala. This pipeline is used to transform translation units into feature vectors and train classifiers using the scikit-learn framework \citep{ScikitLearn}. From the various classification algorithms we tested, Random Forests performed best with our selection of features (see below).

\subsection{Feature Selection}
\label{sec:FeatureSelection}

For the reasons mentioned in \Section{Introduction}, we aimed at finding a combination of features that would perform well with all language pairs rather than tailoring solutions to individual languages. We focused on gearing our classifiers to distinguish correct or almost correct (classes 1, 2) from incorrect TUs (class 3) -- i.e., the Binary Classification (II) task -- by optimising the weighted \F1-score (\F1) on training data (see Tables~\ref{tab:ResultsBinaryTrainingData} and~\ref{tab:ResultsFineGrainedTrainingData}). From the various feature combinations we tested, we found the following to be most successful: \textit{ratio\_words}, \textit{pos\_sim\_all}, \textit{language\_detection}, \textit{mt\_cfs}, \textit{mt\_bleu}, \textit{ratio\_chars} (as described in \Section{Features}), alongside \textit{cg\_score}, \textit{only\_capletters\_dif}, and \textit{punctuation\_similarity} \citep[from][]{Barbu2015}. Evaluation results are given in the next section.

\section{Results}
\label{sec:Results}

We tested our final submission -- a Random Forests classifier based on the nine features described in \Section{FeatureSelection} -- on three language pairs (en--de, en--es, en--it) and two tasks: Binary II and Fine-Grained Classification (see \Sections{BinaryClassification}{FineGrainedClassification}, respectively). The classifier was trained solely on data provided by the organizers of this shared task for each of the language $\times$ task conditions. Each TU in this data was annotated with one of three labels: correct, almost correct, and incorrect (see \Table{Data}).

\subsection{Binary Classification (II)}
\label{sec:BinaryClassification}

Our rationale for focusing on telling apart correct or almost correct from incorrect TUs was that a first application of our method, if successful, would most likely be the filtering of TM data for MT training. While eliminating almost correct TUs might decrease rather than increase MT quality, filtering out incorrect segments can have a positive impact \citep{Vogel2003}.

Prior to submission, we benchmarked our system against the two baselines provided by the organizers: a dummy classifier assigning random classes according to the overall class distribution in the training data (Baseline 1), and a classifier based on the Church-Gale algorithm as adapted by \citet{Barbu2015} (Baseline 2). More importantly, however, we compared our system to \cites{Barbu2015} approach, using the classification algorithms which reportedly worked best with the 17 features in his work. Our system performed well in this comparison, surpassing \citeauthor{Barbu2015}'s approach in all language pairs except en--de, where both systems were \textit{en par}. Details are shown in \Table{ResultsBinaryTrainingData}, where we report weighted precision (P), recall (R), and \F1-scores averaged over 5-fold cross-validation with \nicefrac{2}{3}--\nicefrac{1}{3} splits of the training data.

The final evaluation and ranking produced by the organizers, shown in \Table{ResultsBinaryFinal}, confirms our findings from experimenting with training data: our system performs well on the en--es and en--it test sets (best in class), while performance is substantially lower on the en--de test set. The reasons for this are yet to be ascertained (see also \Section{Conclusions}).

\subsection{Fine-Grained Classification}
\label{sec:FineGrainedClassification}

Although geared to the Binary Classification (II) task (see above), we also assessed our system on the Fine-Grained Classification task. Here, the goal was to distinguish between all of the three classes, i.e., determine whether a TU is correct, almost correct, or incorrect.

Again, we compared our system's performance to \cites{Barbu2015} method, using \nicefrac{2}{3}--\nicefrac{1}{3} splits of the training data (5-fold cross-validation). The results, shown in \Table{ResultsFineGrainedTrainingData}, implied that the nine features we selected would not suffice for a more fine-grained classification of TUs. This was confirmed in the official evaluation and ranking: our system scored low on en--de and mediocre on en--es and en--it. Further work will be needed to analyse these results in more detail.

\section{Conclusions}
\label{sec:Conclusions}

We have proposed a machine learning based method to identify incorrect entries in translation memories. It is applicable to any language pair for which an MT system, a POS tagger, and a language identifier are available. Implemented using off-the-shelf tools, our system achieved the best classification results for two out of three language pairs (English--Italian and English--Spanish) in the Binary Classification (II) task.

In future work, we would like to assess the impact of gearing NLP components to target domains on classification accuracy. The training data in this shared task stems from news (German) and medical texts (Italian, Spanish) which our MT systems, for example, were not optimized for. This domain mismatch might partially explain why our system did not perform well on the English--German test set.

More importantly, however, we would like to test our implementation as-is in Autodesk's production environments for software localization. Removing incorrect segments from TMs could ultimately help make professional translation more efficient by providing better MT (through filtered training data) and more accurate fuzzy matches.

\section{Acknowledgements}

We would like to thank Valéry Jacot for his vital support and guidance.

\section{Bibliographical References}
\label{main:ref}

\renewcommand{\bibsection}{} 
\bibliographystyle{lrec2016}
\bibliography{working-notes}

\end{document}